\documentclass{article} 
\usepackage{iclr2025_conference,times}
\usepackage[utf8]{inputenc}  
\usepackage[french]{babel}
\usepackage{lmodern}         

\usepackage{newtxtext} 

\usepackage{amsmath,amsfonts,bm}

\def\eqref#1{equation~\ref{#1}}

\def\1{\bm{1}}

\DeclareMathAlphabet{\mathsfit}{\encodingdefault}{\sfdefault}{m}{sl}
\SetMathAlphabet{\mathsfit}{bold}{\encodingdefault}{\sfdefault}{bx}{n}

\usepackage{hyperref}
\usepackage{CJKutf8}
\usepackage{tabularx}
\usepackage{booktabs}
\usepackage{makecell}
\usepackage{rotating}
\usepackage{array}
\newcolumntype{L}[1]{>{\raggedright\arraybackslash}p{#1}}

\usepackage{makecell}
\usepackage{tabularx}
\usepackage{booktabs}
\usepackage{makecell}
\usepackage{url}
\usepackage{tabularx} 
\usepackage{booktabs}
\usepackage{graphicx}

\title{MultiNRC: A Challenging and Native Multilingual Reasoning Evaluation Benchmark for LLMs}

\author{Alexander R. Fabbri, Diego Mares, Jorge Flores, Meher Mankikar, \\
\textbf{Ernesto Hernandez, Dean Lee, Bing Liu, Chen Xing} \\
Scale AI
}

\iclrfinalcopy 
\DeclareUnicodeCharacter{202F}{\,}
\begin{document}

\maketitle

\begin{abstract}
Although recent Large Language Models (LLMs) have shown rapid improvement on reasoning benchmarks in English, the evaluation of such LLMs' multilingual reasoning capability across diverse languages and cultural contexts remains limited. 
Existing multilingual reasoning benchmarks are typically constructed by translating existing English reasoning benchmarks, biasing these benchmarks towards reasoning problems with context in English language/cultures. 
In this work, we introduce the Multilingual Native Reasoning Challenge (\textbf{MultiNRC}), a benchmark designed to assess LLMs on more than 1,000 native, linguistic and culturally grounded reasoning questions written by native speakers in French, Spanish, and Chinese.
MultiNRC covers four core reasoning categories: language-specific linguistic reasoning, wordplay \& riddles, cultural/tradition reasoning, and math reasoning with cultural relevance. 
For cultural/tradition reasoning and math reasoning with cultural relevance, we also provide English equivalent translations of the multilingual questions by manual translation from native speakers fluent in English.
This set of English equivalents can provide a direct comparison of LLM reasoning capacity in
other languages vs. English on the same reasoning questions.
We systematically evaluate current 14 leading LLMs covering most LLM families on MultiNRC and its English equivalent set.  
The results show that
(1) current LLMs are still not good at native multilingual reasoning, with none scoring above 50\% on MultiNRC; 
(2) LLMs exhibit distinct strengths and weaknesses in handling linguistic, cultural, and logical reasoning tasks;
(3) Most models perform substantially better in math reasoning in English compared to in original languages (+10\%), indicating persistent challenges with culturally grounded knowledge. 
\end{abstract}

\section{Introduction}
\label{intro}
Enhancing the reasoning capabilities of Large Language Models (LLMs) is a critical area of ongoing research and development~\citep{o3_openai_2024, guo2025deepseek}. 
Consequently, thorough and diverse reasoning evaluation benchmarks in English have been developed to assess LLMs' improved reasoning capabilities. 
These evaluation datasets cover a variety of reasoning tasks, including commonsense reasoning~\citep{zellers2019hellaswag, sakaguchi2021winogrande, clark2018think}, language-understanding reasoning in English~\citep{hendrycks2020measuring, wang2025mmlu}, STEM reasoning~\citep{cobbe2021training, zhong2024agieval, hendrycks2measuring}, and logical reasoning~\citep{liu2020logiqa,yureclor,geva2021did}.
These benchmarks play a critical role in guiding the evolution of the reasoning capacity of LLMs.

While a large number of these benchmarks have been developed in English, the landscape of multilingual reasoning benchmarks is less developed, comprehensive, and equitable.
Existing multilingual reasoning benchmarks are largely created by translating existing English reasoning benchmarks (automatically, manually, or hybrid) into target languages~\citep{shilanguage, patel2021nlp}.
Benchmarks created by directly translating English-centric content may reflect the cultural framing and linguistic characteristic of English, which are not globally representative. 
As a result, reasoning tasks that require native linguistic or cultural understanding are often missing, leaving gaps in the evaluation of non-English language reasoning abilities. 
Consequently, we are still largely unaware of how state-of-the-art (SOTA) LLMs perform on genuinely native multilingual reasoning tasks that demand both linguistic diversity and cultural or contextual depth.

Motivated by this gap, we develop the Multilingual Native Reasoning Challenge (MultiNRC), a new multilingual evaluation benchmark that contains native and challenging reasoning questions of the targeted language, to assess LLMs' reasoning capability under a natural and native context. 
Specifically, MultiNRC contains four categories of reasoning questions: language-specific linguistic reasoning, wordplay riddles, cultural/tradition reasoning, and finally mathematical reasoning with cultural relevance. 
We first release MultiNRC in French, Spanish, and Chinese; Table~\ref{tab:examples_multinrc} shows examples of the four categories in different languages.
To build MultiNRC, we recruit native speakers of each language to create challenging reasoning questions and ground-truth final answers according to the definitions of the four categories above. 
We only accept reasoning questions that cause at least $3$ out of $5$ SOTA LLMs to fail, and we facilitate MultiNRC with automatic evaluation with LLM-as-a-judge, for fast and accurate model assessments. 
Automatic evaluation is made possible by only including reasoning questions with an objective and short ground-truth final answer; we find that our automatic evaluation has more than 95\% alignment with human judgment on MultiNRC.

We thoroughly evaluate current SOTA LLMs on MultiNRC. We find that the best performing LLM, o3-pro~\citep{openai_o3_o4mini_2025} released in April 2025, achieves only 49\% accuracy according to automatic evaluation with respect to our ground-truth final answers, indicating the high difficulty level of MultiNRC for SOTA models.
We find that o3 variants along with Gemini-2.5-Pro (June 2025)~\citep{comanici2025gemini} greatly outperform other models, and that LLMs also rank differently across MultiNRC reasoning categories, suggesting that the taxonomy of MultiNRC provides a more informative view to guide LLM performance improvement.
For example, Gemini-2.5-Pro ranks third on average accuracy among models but ranks first in math reasoning with cultural relevance.

\begin{table}[t!]
\centering
\scriptsize
\begin{tabular}{p{1.2cm} p{1.5cm} p{6cm} p{2.5cm}}
\toprule
\textbf{Language} & \textbf{Category} & \textbf{Prompt} & \textbf{Answer} \\
\midrule
French  & Linguistic   & Si je suis un mot masculin seul, féminin au pluriel, je ne suis jamais bouclé. Qui suis-je? & Délice(s) \\ \hline
Chinese  & Wordplay   & \begin{CJK}{UTF8}{gbsn} 你可以用谐音梗，用一个成语回答我：为什么井越浅越好？因为\_\_\_\_\_\_\_\_\_\_\_\_。 
\end{CJK}  &  \begin{CJK}{UTF8}{gbsn} 好景不长 \end{CJK}  \\ \hline
Spanish  & Cultural/Trad   & Me voy de viaje a Cancún el próximo año en 2026. Quiero estar allá para el Día de la Candelaria, pero quiero tener dos fines de semana completos para disfrutar la playa. ¿Cuáles tendrían que ser las fechas de mis vuelos?  & Ida: viernes, 30 de enero, 2026
Regreso: lunes, 9 de febrero, 2026 \\ \hline
French  & Math   & J'ai trouvé une maison à acheter en viager en 2025 : le bouquet est de 90 000€ et la mensualité de 500€. L'occupante est née l'année de la sixième édition du Festival de Cannes, la maison est estimée à 150 000€, et va prendre 1\% de valeur par an. A partir de quel âge de l'occupante cela devient une moins bonne affaire pour moi?  & 86 ans. \\
\bottomrule
\end{tabular}
\caption{Representative native-language example prompts and short answers from MultiNRC.}
\label{tab:examples_multinrc}
\end{table}

We also augment MultiNRC with English equivalent translations of the non-English prompts by having our annotators translate the prompts into English while retaining the structure and solvability of the original question.
In this English equivalent set, we only translate cultural reasoning and math categories. This is because linguistic and wordplay categories cannot be translated into reasonable equivalents; for example, a play on words in French translated to English would no longer make sense. 
This English-translated evaluation set can provide a direct comparison of LLM reasoning capability in other languages vs. English on same reasoning questions.
We find that LLMs perform much better on the resulting set of English math reasoning, achieving 10\% higher accuracy on average, compared to their performance on the original multilingual math reasoning with cultural relevance.
Our case study shows that the reason for this phenomenon is likely that LLMs are able to better retrieve cultural context/knowledge in English for math problems, compared to retrieving such cultural context/knowledge in its original language. 
However, in the category of cultural reasoning, there is no large performance difference in English equivalent prompts compared to the original multilingual prompts. 
Our case study shows that the cultural/tradition context in this category is much more specific and nuanced compared to the math category, and thus often absent in both the LLM knowledge base of English or original language, leading to no performance improvement on English-equivalent prompts. 
Our contributions are as follows. First, we introduce MultiNRC, a native, challenging multilingual reasoning benchmark authored by native speakers to cover reasoning questions with linguistic and cultural nuances. 
Second, we systematically evaluate 14 strong LLMs in French, Spanish, and Chinese, on both native questions and English translations, to understand the true multilingual reasoning capabilities of models.
Third, we perform extensive ablation studies across languages and reasoning categories, as well as the effect of model translation and reasoning in English to disentangle the impact of language, translation, and cultural context on model reasoning. 
Our analyses highlight the underlying cultural and contextual knowledge required to answer MultiNRC questions.
Finally, we release the dataset to facilitate further research on multilingual and culturally informed evaluation.\footnote{\url{https://huggingface.co/datasets/ScaleAI/MultiNRC}}

\section{Related Work}
\label{related_work}
\textbf{Reasoning benchmarks in English.}  A wide range of benchmarks have been developed to evaluate the reasoning capabilities of LLMs across commonsense, logical, and STEM reasoning domains.
Commonsense reasoning benchmarks aim to evaluate an LLM's ability to apply everyday knowledge and logical inference to understand and interact with the world in a human-like way.  HellaSwag~\citep{zellers2019hellaswag} and WinoGrande~\citep{sakaguchi2021winogrande} are two widely-adopted examples.
Logical and deductive reasoning benchmarks, such as ReClor~\citep{yureclor} and StrategyQA~\citep{geva2021did}, and ARC-AGI~\citep{chollet2024arc}, are designed to evaluate an LLM's capacity to apply formal rules of logic and to perform deductive, inductive, or abductive inference given a set of premises. 
To comprehensively assess both general knowledge and reasoning, multitask benchmarks like MMLU-Pro~\citep{wang2025mmlu} have emerged, expanding on prior efforts in MMLU~\citep{hendrycks2020measuring} by including more rigorous reasoning components and a broader coverage of academic subjects, rather than just factual recall.
STEM and math reasoning benchmarks, such as AIME~\citep{aime2024} and MATH~\citep{hendrycks2measuring}, are specifically designed to assess an LLM's proficiency in solving complex, multi-step mathematical problems.
Distinct from prior benchmarks, we focus on evaluating reasoning over linguistic, wordplay, cultural/tradition, and math questions, each written by native speakers to capture local language and cultural nuance. 

\textbf{Multilingual reasoning benchmarks.} Compared to LLM reasoning evaluation in English, multilingual reasoning benchmarks are less developed. A majority of them are translated from existing English reasoning benchmarks~\citep{li2024eliciting, shilanguage, huang2025benchmax}.
Among them, MMLU-ProX~\citep{mmluprox}, is built by translating the existing English benchmark, MMLU-Pro~\citep{wang2025mmlu}, to evaluate advanced reasoning across multiple languages. The initial translation of the English questions into the other 28 languages was performed using multiple LLMs, followed by an expert review phase.
Similarly, MGSM~\citep{shilanguage} and BenchMAX~\citep{huang2025benchmax} are translated from GSM8K ~\citep{cobbe2021training} and GPQA~\citep{rein2024gpqa}.
BenchMAX is built with an initial machine translation and a meticulous post-editing process in which each sample was independently reviewed by three distinct native-language annotators. 
MGSM was created by manually translating 250 problems selected from the English GSM8K dataset.
An approach to avoid translating existing English benchmarks is by using exam questions in non-English languages~\citep{hardalov2020exams,zhang2023m3exam,wang2024m4u}. However, such benchmarks, with M3Exam as an example, have mostly multichoice questions that assess knowledge instead of reasoning and are too easy for SOTA models, as suggested by their reported GPT-4 results in 2023~\citep{zhang2023m3exam}.
Another example of data collected from native languages is MultiLoko \cite{hupkes2025multiloko}, which collects SQuAD-style \cite{rajpurkar-etal-2016-squad} questions from the most popular Wikipedia pages for each language, rated for their local relevance to the language. While this work also analyzes model performance on English equivalents, the focus of our work is on reasoning questions over linguistic and cultural knowledge and nuance, and not just factual knowledge.

\section{Evaluation Benchmark Overview}
\label{overview}
Below we describe the Multilingual Native Reasoning Challenge (MultiNRC) benchmark, detailing our choice of reasoning categories, English equivalent set of multilingual prompts, as well as dataset construction methodology and statistics.

\subsection{Reasoning Categories}
MultiNRC contains $4$ categories of reasoning questions. 
They are language-specific
linguistic reasoning, wordplay riddles, cultural/tradition reasoning, and finally,
math reasoning with cultural relevance.
We extract and summarize the $4$ categories from preliminary crowd-sourced native and challenging reasoning questions that we collect for multiple languages and cultures.
The $4$ categories are uniquely defined from each other to assess the different capabilities of LLM required in multilingual reasoning tasks. Table~\ref{tab:examples_multinrc} shows example tasks in all categories, which are described in the following paragraphs. Additional examples can be found in the Appendix in Table \ref{tab:additional_examples_multinrc}.

\begin{table}[t!]
\centering
\small
\begin{tabularx}{\textwidth}{l*{5}{>{\centering\arraybackslash}X}}
\toprule
\textbf{Lang} & \textbf{\# Tot.} & \textbf{Linguistic} & \textbf{Wordplay} & \textbf{Cultural/Tradition} & \textbf{Math} \\
\midrule
French   & 338 & 75 & 79 & 107 & 77 \\
Spanish  & 392 & 71 & 93 & 122 & 106 \\
Chinese  & 325 & 71 & 86 & 90 & 78 \\
\midrule
\textbf{Total} & 1055 & 217 & 258 & 319 & 261 \\
\bottomrule
\end{tabularx}
\caption{MultiNRC dataset statistics by language and category.}
\label{tab:dataset_stats}
\end{table}

\textbf{Language-specific Linguistic Reasoning} questions are based on grammatical rules, honorifics, or language-specific conventions that only exist in the target language but not English. Such questions require reasoning about language structure or usage.
Examples under this category include word formation problems, relation inference based on language conventions, etc. 
We enforce that the questions should require multi-step reasoning, instead of mere linguistic feature identification without reasoning. 
In the French example in Table~\ref{tab:examples_multinrc}, the question requires recognizing that the noun délice is masculine in the singular and feminine in the plural, an uncommon grammatical characteristic not present in English.

\textbf{Wordplay \& Riddles} contain puzzles that rely on the clever and ambiguous use of the target language, often involving multiple meanings, homophones, or puns.  
Since such reasoning questions usually use very language-specific homophones or puns, there are no English translations that can provide the exact meaning of such questions. 
Our evaluations later show that this category is one of the most difficult for LLMs.
In the Chinese example shown in Table~\ref{tab:examples_multinrc}, the question requires the respondent to use a Chengyu (Ancient Chinese idiom) involving a pun or homophone as an answer to a playful riddle, thus demanding clever reasoning. 

\textbf{Cultural/Tradition Reasoning} questions require reasoning through timelines, behaviors, or customs derived from local traditions, holidays, or ceremonies.
Similar to other categories, we enforce the assessment of LLM reasoning capabilities and exclude purely factual questions.
In the Spanish example in the table, the question requires identifying the date of a local holiday (Día de la Candelaria) and reasoning about travel dates to ensure two full weekends are spent at the destination, demonstrating an understanding of local traditions and applying multi-step temporal reasoning.

\textbf{Math Reasoning with Cultural Relevance} questions involve calculations based on cultural-specific counting systems, calendars, currencies, or numerical phrasing. As above, it must require at least 1 reasoning step based on a culturally specific element, and we do not allow pure arithmetic or logic without cultural grounding. 
In the French example in the table, the question requires the application of culturally specific knowledge about the 'viager' real estate system and historical events (such as the date of the Cannes Film Festival) to determine the financial break-even point for a property purchase, thereby testing mathematical reasoning within a local context.

\subsection{English Equivalent Prompts}
In order to test the reasoning gap in English vs. non-English languages, we further ask annotators to translate the native prompts and ground-truth final answers into English. 
Annotators judge the validity of machine-translated prompts from 5 SOTA models and choose the most relevant machine-translated prompt from the above pipeline as a base for translation, or translate from scratch as needed.  
Our annotators focus on translating the logic, not literal words, ensuring that the English translation is also culturally grounded and requires similar reasoning.
We only produce English translations for the Cultural/Tradition Reasoning and Math Reasoning categories.
The translated English version retains the structure and solvability of the original, with the translator trying to translate cultural or linguistic characteristics into English.
Examples of English translated prompts and answers are shown in the Appendix in Table \ref{tab:examples_multinrc_english}.
We found that the categories of Language-specific Linguistic Reasoning and Wordplay \& Riddles could not be reliably translated into English, as they often rely on linguistic phenomena that do not translate or exist in English. Thus, we exclude these categories from the English equivalent set.

\subsection{Data Collection/Review and Statistics}

To collect data for MultiNRC, native speakers start by writing reasoning prompts according to the definition of one of the four categories mentioned above. 
For each prompt, the annotator also writes a ground-truth final answer (GTFA), consisting of either a single word or a short phrase/sentence. 
We then pull model responses from five top models at the time of creation across model providers: Google's Gemini-2.5-pro \citep{comanici2025gemini}, o4-mini \citep{openai_o3_o4mini_2025}, Claude 3.7-Sonnet \citep{anthropic2025claude37sonnet}, Deepseek R1 \citep{deepseekai2025deepseekr1incentivizingreasoningcapability}, and Llama-4 Maverick \citep{meta_llama4_2024}. 
The 5 model responses are presented to annotators in a random order.
We only keep the reasoning questions that 3 or more of the 5 models fail to correctly answer.

After the raw data is collected, we have two native-speaker reviewer layers to assess the quality of the data from different aspects. 
The first native-speaker reviewer layer is to assess the quality of the prompt and GTFA pairs. At this layer we would assess both the GTFA's accuracy and the prompt's reasonableness and alignment with category definition.
The second native-speaker reviewer layer assesses whether there are truly 3 out of 5 SOTA model failures, given the verified prompt and GTFA pairs. This is to verify the difficulty level of the collected data. 
We show the statistics of the final dataset after these collection/review steps in Table \ref{tab:dataset_stats}.
We collect roughly the same amount of examples across categories and languages, more than the 250 per-language count of the popular MGSM dataset \citep{shilanguage}.
We also annotate subcategories of the cultural and linguistic nuances present in the prompts. The results of this analysis are found in the Appendix in Table \ref{tab:nuance_clusters}.

\section{Experimental Settings}

\paragraph{Reasoning Settings}
We perform experiments in the following settings:
1) \textbf{Original} in which the model is given the original multilingual question, and prompted to provide an explanation and a short final answer in the native language of the question. 
2) \textbf{En} in which the model answers human-translated English equivalent questions (illustrated in Section 3.2) in English. By comparing \textbf{En} vs \textbf{Original}, we can see the difference in model's abilities to understand and reason about local linguistic and cultural nuance in the local language versus English. \cite{hupkes2025multiloko} refer to this as the Mother Tongue Effect.
3) \textbf{EnT/R} in which the model is prompted to produce an English translation of the native prompt by itself, reason/explain in English and return the native final answer. 
4) \textbf{EnR} in which the model is asked to reason in English on the native prompt and return the native language response. 
The results of these last two settings are found in Appendix \ref{sec:appendix-eval-results}.

\paragraph{Automatic Evaluation} As described above, we collect human judgment of the correctness of model responses for 5 SOTA models on our benchmark. In order to scale annotations, we leverage an LLM judge that takes the prompt, GTFA, and model response and produces a correct/not correct judgment as well as reasoning behind the judgment. We found that GPT-4.1\citep{openai_gpt41_api_2024} as the judge results in an over 95\% agreement with our human judgments and a Scott's Pi \citep{scott1955reliability} of 0.88, indicating strong reliability of our metric, and thus we utilize this automatic judge for all reported results. 
Note that all experimental results use pass@1. 

\paragraph{Model Selection and Inference Parameters} We select 14 model variations across model providers, to obtain as generalizable results as possible given current SOTA models. 
We test a range of recent LLMs including Google's Gemini-2.5-pro \cite{google_gemini_2025}, OpenAI's o3 and o4-mini \cite{openai_o3_o4mini_2025}, gpt-4.1 \cite{openai_gpt41_api_2024}, Anthropic's Claude (Opus, Sonnet, including thinking and non-thinking models), and Llama-4 \cite{meta_llama4_2024}.
Note that -high or -medium refers to the reasoning effort setting.
We generate with max completion tokens of 32k, thinking budget of 16k where applicable, and default to the model provider's default temperature. 

\begin{table}[t!]
\centering
\small
\begin{tabularx}{0.62\textwidth}{l >{\centering\arraybackslash}X}
\toprule
\textbf{Model} & \textbf{Original} \\
\midrule
o3-pro-2025-06-10-high  &  \textbf{49.00 (±3.02)} \\
o3-2025-04-16-high                & \textbf{45.50 (±3.00)} \\
Gemini-2.5-Pro-Preview-06-05      & \textbf{45.12 (±3.00)} \\
o3-2025-04-16-medium                & 44.45 (±3.00) \\
Claude-4-Opus-20250514-thinking   & 33.93 (±2.86) \\
Claude-4-Opus-20250514            & 29.00 (±2.74) \\
Claude-3.7-Sonnet-thinking        & 27.77 (±2.70) \\
Deepseek-R1-0528                  & 27.58 (±2.70) \\
Deepseek-R1                       & 24.27 (±2.59) \\
o4-mini-high                      & 22.18 (±2.51) \\
GPT-4.1                           & 21.23 (±2.47) \\
Claude-4-Sonnet-20250514          & 18.39 (±2.34) \\
GPT-4o                            & 12.42 (±1.99) \\
Llama-4-Maverick                  & 8.44 (±1.68) \\
\bottomrule
\end{tabularx}
\caption{Accuracy (\%) of models in the \textbf{Original} setting (reasoning and answering in the original language) on the full MultiNRC benchmark, with 95\% confidence intervals in parentheses.}
\label{tab:original_full_results}
\end{table}

\begin{table}[t!]
\centering
\small
\resizebox{\textwidth}{!}{
\begin{tabular}{l|ccc|ccc|ccc|ccc}
\toprule
\textbf{Model}
& \multicolumn{3}{c|}{\textbf{Cultural}} 
& \multicolumn{3}{c|}{\textbf{Math}} 
& \multicolumn{3}{c|}{\textbf{Linguistic}} 
& \multicolumn{3}{c}{\textbf{Wordplay}} \\
& \textbf{Fr} & \textbf{Es} & \textbf{Zh}
& \textbf{Fr} & \textbf{Es} & \textbf{Zh}
& \textbf{Fr} & \textbf{Es} & \textbf{Zh}
& \textbf{Fr} & \textbf{Es} & \textbf{Zh} \\
\midrule
o3-pro & 46.7 & \textbf{45.9} & 50.0 & \textbf{57.1} & \textbf{24.5} & 24.4 & \textbf{62.7} & \textbf{47.9} & \textbf{53.5} & \textbf{75.9} & \textbf{53.8} & \textbf{55.8} \\
o3-2025-04-16-high & \textbf{49.5} & \textbf{46.7} & 48.9 & \textbf{50.6} & 23.6 & 24.4 & \textbf{65.3} & \textbf{52.1} & 45.1 & \textbf{67.1} & \textbf{43.0} & \textbf{37.2} \\
Gemini-2.5-Pro-Preview-06-05 & \textbf{52.3} & 40.2 & \textbf{61.1} & 37.7 & \textbf{41.5} & \textbf{43.6} & 52.0 & 39.4 & \textbf{49.3} & 53.2 & 36.6 & 36.0 \\
o3-2025-04-16-medium & \textbf{48.6} & \textbf{46.7} & \textbf{52.2} & \textbf{41.6} & 20.8 & 25.6 & \textbf{66.7} & \textbf{46.5} & 42.3 & \textbf{67.1} & \textbf{37.6} & \textbf{44.2} \\
Claude-4-Opus-20250514-thinking & 36.4 & 20.5 & \textbf{54.4} & 32.5 & 21.7 & \textbf{32.1} & 34.7 & 32.4 & \textbf{46.5} & 48.1 & 32.3 & 25.6 \\
Claude-4-Opus-20250514 & 29.9 & 23.8 & 45.6 & 27.3 & 19.8 & 28.2 & 28.0 & 23.9 & 39.4 & 36.7 & 29.0 & 20.9 \\
Claude-3.7-Sonnet-thinking & 26.2 & 23.8 & 21.1 & 32.5 & \textbf{26.4} & 20.5 & 40.0 & 40.8 & 39.4 & 34.2 & 22.6 & 15.1 \\
Deepseek-R1-0528 & 27.1 & 23.8 & 51.1 & 22.1 & 13.2 & \textbf{30.8} & 25.3 & 42.3 & \textbf{46.5} & 12.7 & 17.2 & 27.9 \\
Deepseek-R1 & 23.4 & 13.9 & 53.3 & 24.7 & 12.3 & 29.5 & 18.7 & 19.7 & 38.0 & 10.1 & 19.4 & 34.9 \\
o4-mini-high & 26.2 & 20.5 & 16.7 & 23.4 & 7.5 & 15.4 & 46.7 & 31.0 & 15.5 & 26.6 & 29.0 & 14.0 \\
GPT-4.1 & 29.0 & 27.0 & 13.3 & 26.0 & 18.9 & 14.1 & 25.3 & 21.1 & 28.2 & 24.1 & 16.1 & 10.5 \\
Claude-4-Sonnet-20250514 & 16.8 & 11.5 & 27.8 & 20.8 & 16.0 & 20.5 & 17.3 & 23.9 & 23.9 & 22.8 & 11.8 & 14.0 \\
GPT-4o & 18.7 & 13.9 & 15.6 & 9.1 & 9.4 & 6.4 & 13.3 & 21.1 & 21.1 & 11.4 & 4.3 & 5.8 \\
Llama-4-Maverick & 13.1 & 8.2 & 5.6 & 11.7 & 6.6 & 10.3 & 9.3 & 12.7 & 11.3 & 8.9 & 2.2 & 3.5 \\
\midrule
\textbf{Average} & \textbf{31.7} & \textbf{26.2} & \textbf{36.9} & \textbf{29.8} & \textbf{18.7} & \textbf{23.3} & \textbf{36.1} & \textbf{32.5} & \textbf{35.7} & \textbf{35.6} & \textbf{25.3} & \textbf{24.7} \\
\bottomrule
\end{tabular}
}
\caption{Model performance in the Original setting broken down by category and language.}
\label{tab:per_lang_per_category}
\end{table}

\begin{table}[h]
\centering
\small
\begin{tabular}{l|ccc|ccc}
\toprule
\textbf{Model}
& \multicolumn{3}{c|}{\textbf{Cultural}}
& \multicolumn{3}{c}{\textbf{Math}} \\
& $\Delta$Fr & $\Delta$Es & $\Delta$Zh & $\Delta$Fr & $\Delta$Es & $\Delta$Zh \\
\midrule
o3-pro & \textbf{-9.3} & 4.1 & 7.8 & -5.2 & \textbf{17.9} & 12.8 \\
o3-2025-04-16-high & -5.6 & -6.6 & \textbf{6.7} & -7.8 & \textbf{16.0} & 10.3 \\
Gemini-2.5-Pro-Preview-06-05 & \textbf{-10.3} & 1.6 & 0.0 & 2.6 & 0.9 & \textbf{10.3} \\
o3-2025-04-16-medium & -4.7 & -1.6 & \textbf{-8.9} & 0.0 & \textbf{22.6} & 10.3 \\
Claude-4-Opus-20250514-thinking & \textbf{-12.1} & 5.7 & -7.8 & 3.9 & \textbf{18.9} & 15.4 \\
Claude-4-Opus-20250514 & -4.7 & 0.8 & \textbf{-3.3} & 3.9 & \textbf{14.2} & 10.3 \\
Claude-3.7-Sonnet-thinking & -0.9 & 0.0 & \textbf{5.6} & -3.9 & \textbf{13.2} & 11.5 \\
Deepseek-R1-0528 & -1.9 & \textbf{17.2} & -2.2 & 9.1 & 33.0 & \textbf{33.3} \\
Deepseek-R1 & \textbf{-6.5} & 4.1 & -7.8 & 0.0 & \textbf{22.6} & 10.3 \\
o4-mini-high & -3.7 & 7.4 & \textbf{8.9} & 1.3 & \textbf{28.3} & 14.1 \\
GPT-4.1 & -0.9 & 1.6 & \textbf{7.8} & 2.6 & \textbf{17.9} & 9.0 \\
Claude-4-Sonnet-20250514 & \textbf{-4.7} & 4.1 & -1.1 & 2.6 & \textbf{13.2} & 6.4 \\
GPT-4o & -1.9 & 4.9 & \textbf{6.7} & 5.2 & \textbf{23.6} & 9.0 \\
Llama-4-Maverick & -2.8 & 5.7 & \textbf{7.8} & 0.0 & \textbf{25.5} & 12.8 \\
\midrule
\textbf{Average} & \textbf{-5.0} & \textbf{3.5} & \textbf{1.4} & \textbf{1.0} & \textbf{19.1} & \textbf{12.5} \\
\bottomrule
\end{tabular}
\caption{Delta (English $-$ Original) in accuracy (\%) for Cultural and Math tasks, according to the Original language. \textbf{Bold} indicates the value with largest magnitude in each category per model. Positive values indicate a performance improvement on English translated queries vs Original.}
\label{tab:delta_cultural_math}
\end{table}

\section{Results and Analysis}
The main results on MultiNRC are found in Table \ref{tab:original_full_results}, showing model performance in the Original setting, along with 95\% confidence intervals.
We note that o3 variants along with Gemini-2.5-pro-preview-06-05 score clearly better than any other model. 
However, no model surpasses 50\% accuracy on the dataset, underscoring the difficulty of MultiNRC and the substantial performance gaps even among top models. 
We also find trends of progress in recent models; for example, Deepseek-R1-0528 improves considerably over the initial Deepseek-R1, and GPT-4.1 outperforms GPT-4o. Furthermore, thinking model variants (e.g., Claude-4-Opus-20250514-thinking) consistently show better performance within the same family, indicating the benefit of enhanced reasoning effort.

In Table \ref{tab:per_lang_per_category}, we break down model performance on MultiNRC in the Original setting by both category and language. 
Math reasoning questions are the most challenging overall, with an average accuracy of just 23.3\% across all questions, and especially low scores in Spanish.
In contrast, linguistic and cultural reasoning tasks see higher scores, with average accuracies of 34.8\% and 31.1\%.
Wordplay remains a consistently difficult category for all languages, although the models score better in French for this category.
Spanish stands out as the most challenging language across categories, particularly for math and wordplay.
Some models exhibit distinct strengths in particular languages and categories, as seen in Deepseek-R1-05-28's higher rank on Chinese Math and Linguistic categories.
These results illustrate that model performance is affected not only by the reasoning category itself but also by the interaction between languages and reasoning categories. 
In Appendix \ref{sec:appendix-eval-results}, we provide an additional comparison of the gap between the top and worst performing language for each model. 

Within categories, we see large performance gaps between models. For example, although top-performing models generally struggle on Math overall, Gemini performs much better than others due to its strong results in Spanish Math questions. Similarly, o3-pro demonstrates a significant improvement in Wordplay, especially French Wordplay (75.9\%). 
As a result, models often rank differently across MultiNRC reasoning categories and languages.
These results suggest that the taxonomy of MultiNRC provides a more informative view to guide LLM performance improvements.

The results described above show a full picture of LLM performance in Original setting of MultiNRC. 
Table \ref{tab:delta_cultural_math} presents the performance delta, calculated as the difference between accuracy on English-translated questions and the original native questions, for both Cultural and Math tasks across French, Spanish, and Chinese.
We observe that scores on English-translated versions of Chinese and Spanish questions are often higher than the original scores, although this is model-dependent, and several top-performing models, such as o3 variants, perform better in French compared to Original.
These results suggest that for certain languages and categories, language-specific strengths and weaknesses remain a significant factor in LLM performance. 

For math reasoning, we observe a substantial improvement when models are given English equivalents, with especially pronounced gains for Spanish (+19.1) and Chinese (+12.5). 
Our case studies highlight that LLMs are able to surface the correct background knowledge and relevant values more reliably for the math problems presented in English, leading to higher accuracy on these tasks.
In contrast, for cultural reasoning, the difference is negligible on average.
This suggests that while translation to English can unlock knowledge for certain tasks such as math reasoning questions, performance improvement on culturally grounded tasks remains limited, likely due to the lack of exposure of the models to specific cultural knowledge, regardless of language.
We present case studies of examples where o3 answers English equivalents correctly but provides incorrect answers to the original questions in Appendix~\ref{sec:appendix-eval-results}, along with more quantified analysis on model discrepancies and consistency.

\begin{table}[t!]
\centering
\scriptsize
\begin{tabular}{p{4.4cm} p{1.8cm} p{3.1cm} p{3.1cm}}
\toprule
\textbf{Prompt} & \textbf{Gold Answer} & \textbf{Gemini Response} & \textbf{o3 Response} \\
\midrule
En un documento antiguo se hablaba de que la cosecha de uva en el Ribeiro fue de 50 codos de ribera cúbicos y que en la bodega había 11 bocoys de 40 arrobas. Si el rendimiento de la uva es de un 75\%. Según la correspondencia de medidas recogida en Wikipedia, tomando 3 decimales para la arroba y 5 para el codo de ribera, ¿cuántos porrones de litro y medio hacen falta para guardar el vino que sobra? & 13 & 15 porrones & 9 872 porrones \\
\midrule
Tu dois trouver dans cette conversation la subtilité française:
-J'en ai marre du chantier, j'ai assez donné aujourd'hui.
-Pareil, j'en peux plus...
-On en a marre tôt on vient juste de commencer.
-Tu as raison, qu'est-ce qu'on pourrait faire pour se motiver entre nous ?
-Juste siffloter et attendre la fin de la journée. & marre tôt $->$ marteau  &   La subtilité est la proposition d'une solution passive et résignée ("siffloter et attendre") à un problème qui appelle une action motivante.  &  le jeu de mots « marre tôt » / « marteau » \\ \midrule
\begin{CJK}{UTF8}{gbsn} 快到中秋节了，家里已经准备好月饼和茶，期待当天跟家人一起赏月和提灯笼。今年是2020年后的第一个无春年，早上看了划龙舟比赛，下午还吃了肉粽。对了，从明天算起距离中秋节还有几天？ \end{CJK} & \begin{CJK}{UTF8}{gbsn} 99天  \end{CJK} &   \begin{CJK}{UTF8}{gbsn}  99天  \end{CJK} &  \begin{CJK}{UTF8}{gbsn}   "今年是2020年后的第一个无春年"指的是...（2021 年）。
...今天是端午节...2021 年端午节...是 6 月 14 日，所以"明天"是 2021-06-15。
2021 年中秋节...是 2021-09-21。
计算...天数差：6 月剩余 15 天（15~30 日）...7 月 31 天...8 月 31 天...9 月前 21 天...15 + 31 + 31 + 21 = 98 天。 \end{CJK} \\
\bottomrule
\end{tabular}
\caption{Representative native-language prompts and model outputs for o3 and Gemini models, illustrating both models' strengths and weaknesses. For the top example, both models fail to correctly retrieve the codo de ribera unit. In the second example, Gemini struggles to understand the French wordplay, while o3 incorrectly performs date calculations in the third row.}
\label{tab:rep_errors}
\end{table}

\section{Discussion and Limitations}
We analyze the errors of o3-2025-04-high and Gemini models and include representative examples in Table \ref{tab:rep_errors}. 
We observe that 66\% of the problems are either solved correctly or incorrectly by both models.
Among the problems that the two models perform differently on, our error analysis suggests that o3 performs better on French Wordplay examples, where Gemini is more likely to miss cultural nuance or wordplay (yielding literal or surface-level responses). 
However, Gemini performs better on Chinese cultural prompts. 
The Spanish example in Table \ref{tab:rep_errors} is an example in which both models fail. 
Both models used incorrect values for the codo de ribera unit, leading to a difference in calculating the number of porrones required.
In the French example shown, Gemini fails to recognize the homophonic pun between "marre tôt" and "marteau," while o3 correctly identifies and explains this wordplay. In the Chinese example, o3 undercounts the days between festivals by one.

These case studies highlight the sensitivity of large language models to linguistic and cultural nuances, especially when reasoning in languages other than English.
In our work, we do not explore models designed or finetuned specifically to be multilingual, and we leave an evaluation of these models on the benchmark for future work.
We selected French, Spanish, and Chinese as initial languages given the large number of global speakers and the authors' familiarity with these languages.
To draw wider conclusions about multilingual performance, it will be be important to include additional languages in future work. 
We expect that extending this benchmark to low-resource languages would magnify the struggle of current models in dealing with local knowledge, linguistic phenomena, and complex reasoning. 
Furthermore, our analysis does not explicitly consider the effect of regional or dialectal variation, which is present in languages like Spanish and Chinese, and future work could examine how these differences impact model performance.
This underscores the need for both more diverse training data and tailored model evaluation to ensure robust, equitable progress in multilingual LLMs.
\section{Conclusion}

We introduce MultiNRC, a challenging benchmark designed to evaluate the multilingual abilities of LLMs through linguistically and culturally-nuanced reasoning questions.
By constructing diverse native reasoning questions in French, Spanish, and Chinese, we highlight ongoing challenges for LLMs when faced with linguistically and culturally nuanced reasoning tasks across multiple languages.
Our evaluation shows that while English translation can boost performance on some tasks, substantial gaps remain for many culturally and linguistically nuanced questions.
We hope that MultiNRC will serve as a robust testbed for future advances in multilingual LLM development, and we release our dataset to further research in this direction.

\bibliography{iclr2025_conference}
\bibliographystyle{iclr2025_conference}

\appendix
\section{Appendix}

\subsection{Dataset Examples and Linguistic Phenomena}
\label{sec:appendix-dataset-examples}
\paragraph{Additional Samples}
Table~\ref{tab:additional_examples_multinrc} includes a broader sample of MultiNRC questions. 
For example, one French cultural question requires knowledge of both local festivals and the French calendar of saints. The French wordplay question relies on recognizing that "mon chien Michel" is a homophone for "Mont Saint-Michel," combining everyday language with a cultural reference.

We also include English-equivalent prompts in Table \ref{tab:examples_multinrc_english}.
For example, the French mathematical question asks for the difference in height between Paris's iconic monument and the tallest residential tower in the country, which are translated from French while keeping the logic and required knowledge for the question intact.

\paragraph{Cultural and Linguistic Nuances}
In order to analyze the presence of cultural and linguistic nuances in our prompts, we create categories of linguistic and cultural nuance using an LLM. Then, we label each prompt in our dataset with its associated nuances. The results of the most common linguistic and cultural phenomena present in the dataset are shown in Table \ref{tab:nuance_clusters}.
The categories of Puns, Homophones, \& Wordplay are very common as this includes all examples in the Wordplay category as well as many under Linguistic reasoning. 
Similarly, Calendar \& Time Systems are often present in Math as well as Cultural/Traditions reasoning problems.

\subsection{Additional Evaluation Results}
\label{sec:appendix-eval-results}

\paragraph{Language Consistency and Gaps}

As in \cite{hupkes2025multiloko}, we show the Mother Tongue Effect, calculated as the average performance in the original language minus the average performance in English over all questions. 
However, in contrast to \cite{hupkes2025multiloko}, here we note a negative effect on linguistically and culturally nuanced reasoning questions. 
We also include model consistency, as a measure of the percentage of questions that a model answers correctly in either original or En divided by the number of questions the model answers correctly in total. 
For most models, the consistency is around or below 50\%, but we note that the top-performing models also have higher consistency, and that consistency has a strong rank correlation with original. 
The GAP column quantifies this difference between the best-performing language and the worst for each model, with an average gap of 13.90 points across all models.
We observed that some models show especially large differences across languages; for example, Deepseek-R1-0528 scores 22.81 points higher in Chinese than in Spanish, and the original Deepseek-R1 a 29.10 difference.
This underscores that even strong multilingual models perform very differently across languages.

\paragraph{Original vs En Examples}
We provide two examples where o3-2025-04-16-high incorrectly answers the question in French while correctly responding to the English translation in Table \ref{tab:incorrect_original_correct_en}.
In the first example, o3 misinterprets the value of a coin in French but correctly identifies it and computes and resulting answer in English. In the second example, a riddle about the color of a sash, o3 incorrectly determines the color sequence and predicts the wrong gender, while in English the model correctly understands the intended colors and predicts the right answer. 
These cases illustrate how the model may better surface and reason over certain cultural information in English, leading to improved performance. 

\paragraph{Machine Translation and English Reasoning}
When comparing the reasoning settings in Table \ref{tab:main_results}, we find a slight improvement with EnR on average, but this effect varies depending on the model. 
Furthermore, EnT/R does not offer an improvement on average. The combination of translation error due to native prompt misunderstanding likely contributes to error propagation.
We show the delta between Original and En/R according to categories and languages in Table \ref{tab:delta_all_categories_enreason}.

We note some improvement from reasoning in English for Chinese prompts, with Math, Linguistic, and Wordplay categories showing average deltas above +1.7. However, we see no improvements in the Cultural category.
Even within a particular model, the benefit of reasoning in English often varies depending on both category and language. 
These results highlight the inherent differences in model families and the heterogenous impact of translation and English reasoning, reinforcing the need for language-specific and model-specific analysis in multilingual benchmarks. .

\begin{table}[t!]
\centering
\scriptsize
\begin{tabular}{p{1.2cm} p{1.5cm} p{6cm} p{2.5cm}}
\toprule
\textbf{Language} & \textbf{Category} & \textbf{Prompt} & \textbf{Answer} \\
\midrule
Chinese  & Math    &  \begin{CJK}{UTF8}{gbsn} 我正在写小说，这是小说里的一个场景：我参加外婆的寿宴时，外婆很自豪地说：“算起来，这年是我第九轮的第一年。我可是民国二十年生人。”。我妈说：”妈，时间过得真快，我也年过花甲了。“这时小姨插嘴道：”二姐，我比你小一岁。我是属马的。“我知道这里提到的都是虚岁，这时候我弟弟就问我，外婆，妈妈和小姨的到底实岁是几岁？外婆和妈妈相差几岁？ \end{CJK} & 1 \\ \hline
Spanish  & Math   &  Mi jefa quedó engatusada con la oferta de un vendedor de refris en Tijuana, y quiero hacer que agarre el rollo. Se me hace que se la quieren fregar, pero nomás no le cae el veinte. Échame la mano con los números.

El refri cuesta 20 milagritos ya con IVA. También que su pago mensual por 12 meses será igual al pago original del IVA. Este bato dice que le van a regresar el IVA del precio final que haya pagado.

¿Cuál es el resultado del total que pagó por el refri menos lo que le regresen?  & \$16,355.56 \\ \hline
Chinese  & Math   &  \begin{CJK}{UTF8}{gbsn} 我最小的妹妹是2010年后出生的，出生在一个双春年的中秋节。我们家非常重视中秋节，所以虽然她出生证明上的日期是阳历生日，但她的生日派对每年都在中秋节这一天举行的。
今年是闰六月年份，根据阳历计算，今年的中秋节比她出生那天的中秋节晚了一个数字 X，这个数字 X 刚好等于今年和她出生年份之间的相差年数。请问，我妹妹出生于几年几月几号，今年几岁了？\end{CJK} & \begin{CJK}{UTF8}{gbsn} 2020年10月1日；5岁 \end{CJK} \\ \hline
Chinese  & Cultural    & \begin{CJK}{UTF8}{gbsn} 我姐夫的家乡在潮汕，当时他和姐姐的婚礼定在了当地俗称“天日”的这天，希望可以带来好兆头。我依稀记得他们结婚那年是长辈说的辛丑年，如果我和男朋友2024年也在相同的公历日期结婚，你知道确切的日子吗？ \end{CJK} & \begin{CJK}{UTF8}{gbsn} 2024年2月20日 \end{CJK} \\ \hline
French  & Cultural     &  En 2025, comme tous les ans, les Fêtes de la Madeleine se dérouleront à Mont-de-Marsan. Quel prénom est célébré le dernier jour des Fêtes de la Madeleine selon le calendrier des saints en France ?  & Marina \\ \hline
French  & Wordplay   & Explique cette devinette  : Pourquoi un Normand et un Breton se disputent-ils ? Pour savoir à qui appartient mon chien Michel.   & "mon chien Michel" sonne comme "Mont Saint-Michel". \\ \hline
Spanish  & Wordplay   &  Mi compañera estaba riendo feliz mientras yo, tristemente, sacudía los muebles. ¿Qué hace mi compañera?  & Barriendo. \\ \hline
Spanish  & Linguistic   & Estoy observando desde la puerta de una casa abandonada a la noche algunas cosas que están a mi lado, otras a unos pasos más alejadas y otros a varios metros de distancia: 

Aquello está un poco marchito. 
Esta cosa le falta un ojo. 
Esos tienen un mango
Aquella contiene múltiples herramientas. 
Aquí tengo un juguete que tiene ruedas. 

Al día siguiente, con mayor iluminación volví a observar desde la puerta qué elementos hay. 

Esto es un tren
Allá están las flores. 
Ahí está el coso para golpear. 
Eso es para cavar.
Al lado mío está un muñeco de trapo. 
Allá está la caja.

Al tercer día, un fantasma movio las cosas: 

Aquello lo trajo acá. 
Lo de ahí fue allá. 

Indicar cuantos objetos hay acá, ahí y allá.  &   4 acá, 0 ahí y 2 allá. \\ 
\bottomrule
\end{tabular}
\caption{Additional native-language example and ground-truth answers from MultiNRC.}
\label{tab:additional_examples_multinrc}
\end{table}

\begin{table*}[t]
\centering
\small
\setlength{\tabcolsep}{6pt}
\renewcommand{\arraystretch}{1.12}
\resizebox{\textwidth}{!}{
\begin{tabular}{p{1.2cm} p{1.6cm} p{3.5cm} p{1.1cm} p{3.5cm} p{3.5cm}}
\toprule
\textbf{Language} & \textbf{Category} & \textbf{Prompt} & \textbf{Answer} & \textbf{English Prompt} & \textbf{English GTFA} \\
\midrule
French & Cultural Reasoning & Je suis agent chargé de la tranquillité publique à Coin-sur-Seille et il me reste 4 jours de congés à poser en 2025. J'aimerais profiter de toute la semaine qui inclut le Vendredi saint. Est-ce que c'est envisageable avec le solde qui me reste ? & Non. & I work as a public safety officer in Coin-sur-Seille and only have 4 days of leave left to take. Id like to take off the entire week that includes Good Friday. Is this possible with my remaining balance? & No, this is not an option. \\
\midrule
French & Math & Quelle est la différence entre la hauteur du monument emblématique de Paris et la plus haute tour habitable du pays? & 99 mètres & What is the difference between the height of Paris' iconic monument and the tallest residential tower in the country? & The difference between the two buildings is 99 meters \\
\midrule
Spanish & Cultural Reasoning & Mi abuelo nació justamente 10 años antes de que Argentina retomara las Islas Malvinas, yo nací 30 años después el Día del Veterano y de los Caídos en la guerra de Malvinas ¿En qué día de la semana nací yo? & Martes & My grandfather was born exactly 10 years before Argentina recovered the Malvinas (Falkland) Islands, and I was born 30 years later on the Day of the Veteran and Fallen in the Malvinas War (Día de los Veteranos y Caicos in las Guerra de Malvinas). On what day of the week was I born? &  Tuesday \\
\midrule
Spanish & Math & Mis compas y yo somos futboleros de corazón. No hay partido de la Selección Nacional que dejemos de ver. Uno de mis amigos, Carlos, es 3 años menor que yo. El otro, Santi, es 2 años menor que Carlos y Fer, es 1 año mayor que yo. Si yo nací 18 años antes del segundo año de la temporada de Miguel Mejía Barón, como entrenador de la Selección, ¿cuántos años tenía Santi en julio del año pasado, 2024, si todos nacimos en el mes de agosto? & 42 años. & My friends and I are die-hard soccer fans. There isn’t a match of the National Team that we miss watching. One of my friends, Carlos, is three years younger than me. The other one, Santi, is two years younger than Carlos, and Fer is one year older than me. If I was born eighteen years before the second year of Miguel Mejía Barón’s tenure as coach of the National Team, how old was Santi in July of last year, 2024, if we were all born in August? & 42 years old. \\
\midrule
Chinese & Cultural Reasoning & \begin{CJK}{UTF8}{gbsn} 有一位清朝皇帝是首位出生和成长在北京的皇帝。请问他的最后一位皇后当了多久皇后就过世了？ \end{CJK} & \begin{CJK}{UTF8}{gbsn} 1天 \end{CJK} & There was a Qing dynasty emperor who was the first to be born and raised in Beijing. How long did his last empress serve as empress before she died? & One day. \\
\midrule
Chinese & Math & \begin{CJK}{UTF8}{gbsn} 据说关羽身高9尺的青龙偃月刀重82斤，请问是现在的多高，刀又有多少公斤？ \end{CJK} & \begin{CJK}{UTF8}{gbsn} 217厘米（据找到的资料一尺一般在23.1厘米到23.3厘米之间），18.04公斤 \end{CJK} & It is said that Guan Yu was 9 chi tall and his Green Dragon Crescent Blade weighed 82 jin. Could you tell me how tall that is in modern measurements, and how many kilograms the blade weighed? & Guan Yu was approximately 2.08 meters tall, and his Green Dragon Crescent Blade weighed approximately 18.04 kilograms. \\
\bottomrule
\end{tabular}
}
\caption{Representative native-language prompts and answers from MultiNRC, with English prompt and reference answer.}
\label{tab:examples_multinrc_english}
\end{table*}

\begin{table}[t!]
\centering
\small
\begin{tabularx}{\textwidth}{X r | X r}
\toprule
\multicolumn{2}{c|}{\textbf{Linguistic Nuances}} & \multicolumn{2}{c}{\textbf{Cultural Nuances}} \\
\textbf{Category} & \textbf{Count} & \textbf{Category} & \textbf{Count} \\
\midrule
Puns, Homophones \& Wordplay             & 454  & Calendar \& Time Systems                    & 324 \\
Polysemy, Homonymy \& Ambiguity          & 317  & Specific Units \& Currency  & 222 \\
Discourse Inference \& Pragmatics        & 167  & Kinship \& Social Hierarchies               & 128 \\
Slang \& Colloquial Language             & 165  & Cross-System or Interlingual Reasoning      & 107 \\
Idioms, Proverbs \& Figurative Language  & 139  & Institutional, Administrative, or Legal & 80 \\
Lexical Gaps or Untranslatables          & 126  & Embedded or Hidden-Form Solutions           & 51 \\
Gendered Language \& Agreement           & 58   & Game Rules \& Contextual Calculation        & 42 \\
Formal Register \& Politeness            & 31   &                                            &     \\
\bottomrule
\end{tabularx}
\caption{Clusters of top categories of linguistic and cultural nuances present in MultiNRC prompts.}
\label{tab:nuance_clusters}
\end{table}

\begin{table}[t!]
\centering
\small
\begin{tabular}{l|r|r|r}
\toprule
\textbf{Model} & \textbf{MTE} & \textbf{Consistency} & \textbf{GAP} \\
\midrule
o3-pro & -4.66 & 61.98 & 15.12 \\
o3-2025-04-16-high & -1.90 & 52.04 & 14.04 \\
Gemini-2.5-Pro-Preview-06-05 & -0.34 & 54.47 & 12.19 \\
o3-2025-04-16-medium & -2.93 & 47.68 & 11.00 \\
Claude-4-Opus-20250514-thinking & -3.79 & 37.28 & 22.99 \\
Claude-4-Opus-20250514 & -3.28 & 33.46 & 15.57 \\
Claude-3.7-Sonnet-thinking & -4.14 & 34.19 & 7.97 \\
Deepseek-R1-0528 & -14.66 & 30.42 & 22.81 \\
Deepseek-R1 & -3.97 & 31.51 & 29.10 \\
o4-mini-high & -9.48 & 27.75 & 10.53 \\
GPT-4.1 & -6.21 & 31.82 & 14.03 \\
Claude-4-Sonnet-20250514 & -3.45 & 23.40 & 10.81 \\
GPT-4o & -7.93 & 17.79 & 3.36 \\
Llama-4-Maverick & -8.28 & 12.41 & 5.04 \\
\midrule
\textbf{Average} & -5.36 & 35.44 & 13.90 \\
\bottomrule
\end{tabular}
\caption{
MTE (Mother Tongue Effect: average difference Original $-$ English), Consistency (proportion correct in both settings), and GAP (difference between highest and lowest original language score) for each model. Note that GAP is calculated over all datapoints, while MTE and Consistency over those with English translations.
}
\label{tab:mte_consistency_gap}
\end{table}

\begin{table*}[h]
\centering
\small
\resizebox{1\textwidth}{!}{
\begin{tabular}{p{4.5cm} p{2cm} p{4.5cm} p{3cm} p{2.8cm}}
\toprule
 \textbf{Prompt} & \textbf{GTFA} & \textbf{English Prompt} & \textbf{o3 original Answer} & \textbf{o3 English Answer} \\
\midrule
 Je crois que j'ai perdu de l'argent de ma poche aujourd'hui. Je me rappelle qu'avant de sortir j'avais 4 billets verts et une pièce argentée avec contour doré, mais en rentrant j'ai trouvé dans ma poche uniquement la somme de 200. Quelle somme aurais-je perdu ? & 201 euros & I think I lost some money from my pocket today. I remember that before I went out, I had 4 green bills and one silver coin with a golden contour, but when I came back, I only found 200 in my pocket. How much money would I have lost? &
\textbf{Explication:} ...
La « pièce argentée avec contour doré » correspond à la pièce bicolore de 2 €.   ... \textbf{Exact Answer:} 202~€ &
\textbf{English Explanation:} In euro currency: \newline
• The only bill that is predominantly green is the €100 note....\newline
• A coin that is silver-coloured ... with a golden ring around it matches the €1 coin. \newline
\textbf{Exact Answer:} €201
\\
\midrule

Ma femme est allée seule à l'échographie de notre second enfant permettant de savoir si c'est un garçon ou une fille...Je n'étais pas disponible. Elle a choisi de me le faire deviner sous forme d'une énigme basée sur les couleurs :

1. Je suis la couleur la plus proche du col de l'écharpe tricolore d'un parlementaire lorsqu'elle est portée en écharpe.
2. Je suis la couleur du ruban de l'insigne d'un chevalier de la Légion d'honneur.
3. Si les couleurs sont différentes, alors c'est un garçon. Si elles sont identiques, alors c'est une fille.

Alors, verdict ? & Une fille (a girl) & My wife went alone to the ultrasound of our second child to find out if it's a boy or a girl... I wasn't available. She chose to have me guess in the form of a riddle based on colors:

1. I am the color closest to the collar of the tricolor sash of a parliamentarian when it is worn as a sash.
2. I am the color of the ribbon of the insignia of a Knight of the Legion of Honor.
3. If the colors are different, then it's a boy. If they are identical, then it's a girl.

So, verdict? &
... l'ordre des bandes est bleu-blanc-rouge ; la couleur la plus proche du col est donc le bleu.
Le ruban de l'insigne de chevalier de la Légion d'honneur est entièrement rouge.  
Comme les deux couleurs obtenues ... ne sont pas identiques, la règle donnée ...  le bébé est un garçon.

 &
1. When a French parliamentarian ... wears the tricolour sash, the stripe that sits nearest the collar is red.
2. The ribbon from which the insignia of a Knight (Chevalier) of the Légion d'honneur is suspended is entirely red.
Both requested colours are therefore red.  
Because your wife's rule ... the identical reds mean the baby is a girl.

\\
\bottomrule
\end{tabular}
}
\caption{Examples where o3 incorrectly answers the French prompt but correctly answer the English-translated problem.}
\label{tab:incorrect_original_correct_en}
\end{table*}

\begin{table}[t!]
\centering
\small
\begin{tabularx}{\textwidth}{l *{3}{>{\centering\arraybackslash}X}}
\toprule
\textbf{Model} & \textbf{Orig} & \textbf{EnT/R} & \textbf{EnR} \\
\midrule
o3-pro-2025-06-10-high                  & \textbf{49.00} & 41.71 & \textbf{45.97} \\
o3-2025-04-16-high                 & \textbf{45.50} & \textbf{45.02} & \textbf{46.54} \\
Gemini-2.5-Pro-Preview-06-05       & 45.12 & 44.08 & 45.50 \\
o3-2025-04-16-medium         & 44.45 & \textbf{45.21} & 45.31 \\
Claude-4-Opus-20250514-thinking    & 33.93 & 32.99 & 33.27 \\
Claude-4-Opus-20250514             & 29.00 & 26.26 & 26.82 \\
Claude-3.7-Sonnet-thinking         & 27.77  & 25.02 & 26.54 \\
Deepseek-R1-0528                   & 27.58 & 24.36 & 28.91 \\
Deepseek-R1                        & 24.27 & 21.99 & 24.17 \\
o4-mini-high                       & 22.18 & 24.64 & 25.21 \\
GPT-4.1                            & 21.23 & 22.65 & 22.09 \\
Claude-4-Sonnet-20250514           & 18.39 & 16.68 & 17.91 \\
GPT-4o                             & 12.42 & 12.23 & 11.56 \\
Llama-4-Maverick                   & 8.44  & 9.48  & 10.81 \\
\midrule
\textbf{Average}                   &   31.45 &  30.38 & 31.67 \\
\bottomrule
\end{tabularx}
\caption{Accuracy (\%) of models under different multilingual reasoning settings: \textbf{Original} (reasoning in the original language), \textbf{EnT/R} (translate prompt to English, then reason in English), and \textbf{EnR} (reasoning in English).}
\label{tab:main_results}
\end{table}

\begin{table}[h]
\centering
\small
\resizebox{\textwidth}{!}{
\begin{tabular}{l|ccc|ccc|ccc|ccc}
\toprule
\textbf{Model}
& \multicolumn{3}{c|}{\textbf{Cultural}}
& \multicolumn{3}{c|}{\textbf{Math}}
& \multicolumn{3}{c|}{\textbf{Linguistic}}
& \multicolumn{3}{c}{\textbf{Wordplay}} \\
& $\Delta$Fr & $\Delta$Es & $\Delta$Zh
& $\Delta$Fr & $\Delta$Es & $\Delta$Zh
& $\Delta$Fr & $\Delta$Es & $\Delta$Zh
& $\Delta$Fr & $\Delta$Es & $\Delta$Zh \\
\midrule
o3-pro & 0.9 & -1.6 & \textbf{-4.4} & \textbf{-6.5} & -0.9 & 0.0 & \textbf{-10.7} & -7.0 & -4.2 & -2.5 & \textbf{-5.4} & 2.3 \\
o3-2025-04-16-high & -0.9 & 0.8 & \textbf{5.6} & \textbf{-6.5} & 3.8 & 5.1 & 4.0 & \textbf{-5.6} & \textbf{5.6} & 0.0 & -7.5 & \textbf{8.1} \\
Gemini-2.5-Pro-Preview-06-05 & 0.0 & 2.5 & \textbf{-4.4} & 5.2 & \textbf{-7.5} & 1.3 & 1.3 & \textbf{2.8} & 1.4 & -1.3 & 2.2 & \textbf{3.5} \\
o3-2025-04-16-medium & 2.8 & -1.6 & \textbf{-8.9} & 3.9 & \textbf{4.7} & 1.3 & 2.7 & 1.4 & \textbf{11.3} & \textbf{-3.8} & 0.0 & -1.2 \\
Claude-4-Opus-20250514-thinking & 2.8 & -1.6 & \textbf{-3.3} & 3.9 & \textbf{5.7} & 1.3 & 4.0 & -5.6 & \textbf{-7.0} & \textbf{-7.6} & -3.2 & 0.0 \\
Claude-4-Opus-20250514 & -3.7 & \textbf{-9.8} & -1.1 & \textbf{-5.2} & 0.0 & -3.8 & \textbf{-6.7} & 4.2 & 4.2 & \textbf{2.5} & -2.2 & 0.0 \\
Claude-3.7-Sonnet-thinking & -4.7 & -2.5 & \textbf{6.7} & -2.6 & -3.8 & \textbf{5.1} & \textbf{-13.3} & 1.4 & 2.8 & \textbf{2.5} & -4.3 & 0.0 \\
Deepseek-R1-0528 & 0.9 & \textbf{4.1} & -3.3 & \textbf{7.8} & 1.9 & 3.8 & -2.7 & \textbf{-5.6} & 2.8 & \textbf{8.9} & -1.1 & -2.3 \\
Deepseek-R1 & -2.8 & -2.5 & \textbf{-5.6} & -6.5 & 0.9 & \textbf{7.7} & 5.3 & \textbf{9.9} & -7.0 & \textbf{5.1} & -3.2 & 1.2 \\
o4-mini-high & 3.7 & 2.5 & \textbf{5.6} & \textbf{2.6} & 0.0 & 1.3 & 0.0 & -1.4 & \textbf{5.6} & \textbf{12.7} & -5.4 & 10.5 \\
GPT-4.1 & 1.9 & -0.8 & \textbf{3.3} & -1.3 & \textbf{2.8} & -1.3 & \textbf{-5.3} & \textbf{8.5} & 4.2 & \textbf{-6.3} & 1.1 & 3.5 \\
Claude-4-Sonnet-20250514 & -0.9 & \textbf{-1.6} & -1.1 & 0.0 & \textbf{-8.5} & -2.6 & 0.0 & 4.2 & \textbf{5.6} & -3.8 & \textbf{5.4} & 1.2 \\
GPT-4o & 0.0 & 2.5 & \textbf{-5.6} & \textbf{3.9} & 0.0 & 1.3 & \textbf{-4.0} & 0.0 & -2.8 & -3.8 & 1.1 & \textbf{-4.7} \\
Llama-4-Maverick & 0.0 & -2.5 & \textbf{8.9} & 1.3 & 0.9 & \textbf{5.1} & 0.0 & 4.2 & \textbf{5.6} & 0.0 & \textbf{5.4} & 2.3 \\
\midrule
\textbf{Average} & \textbf{0.0} & \textbf{-0.9} & \textbf{-0.6} & \textbf{0.0} & \textbf{0.0} & \textbf{1.8} & \textbf{-1.8} & \textbf{0.8} & \textbf{2.0} & \textbf{0.2} & \textbf{-1.2} & \textbf{1.7} \\
\bottomrule
\end{tabular}
}
\caption{Delta (EnR $-$ Original) in accuracy (\%) for each reasoning category and language. \textbf{Bold} indicates the value with largest magnitude in each category per model. Positive values indicate a performance improvement on queries when reasoning in English vs Original.}
\label{tab:delta_all_categories_enreason}
\end{table}

\end{document}